\documentclass[journal]{IEEEtran}

\usepackage{dblfloatfix}
\usepackage[T1]{fontenc}
\usepackage{multirow}
\usepackage[table,xcdraw]{xcolor}
\usepackage{graphicx}
\usepackage{amsmath}
\usepackage{amsfonts}
\usepackage{algorithm}
\usepackage{amsthm,amssymb}
\usepackage{mathrsfs}
\usepackage{algorithmic}
\usepackage{makecell}
\usepackage{verbatim}
\usepackage{booktabs}
\setcellgapes{3pt}
\usepackage{amsthm,amsmath,amssymb,lipsum}
\usepackage{mathrsfs}
\setcellgapes{3pt}
\usepackage{multirow}

\usepackage[justification=centering]{caption}
\usepackage{color}
\usepackage[subfigure]{tocloft}
\usepackage{subfigure}
\usepackage{tabularx}
\usepackage{amsmath, bm}
\usepackage{authblk}
\usepackage{amsmath}
\usepackage{multirow}
\usepackage{tikz}
\usetikzlibrary{shapes.geometric, arrows, positioning, fit, calc}
\usepackage{bm}
\makeatletter

\newcommand{\Rmnum}[1]{\expandafter\@slowromancap\romannumeral #1@}
\makeatother

\usepackage{mdframed}
\usepackage{listings}
\usepackage{xcolor}

\mdfdefinestyle{promptbox}{
    linecolor=black,
    linewidth=1pt,
    frametitlerule=true,
    frametitlebackgroundcolor=gray!20,
    innertopmargin=10pt,
    innerbottommargin=10pt,
    innerrightmargin=10pt,
    innerleftmargin=10pt,
    skipabove=10pt,
    skipbelow=10pt
}

\usepackage{hyperref} 
\hypersetup{
	hidelinks
}
\ifCLASSINFOpdf
\else
\fi
\usepackage{hyperref}

\begin{document}

\title{Agentic AI Empowered Intent-Based Networking for 6G}

\author{
Genze Jiang, Kezhi Wang, \textit{Senior Member, IEEE}, Xiaomin Chen, Yizhou Huang
	\thanks{
    Genze Jiang, Kezhi Wang and Yizhou Huang are with the Department of Computer Science, Brunel University London, UK (e-mail: genze.jiang@brunel.ac.uk, kezhi.wang@brunel.ac.uk, yizhou.huang2@brunel.ac.uk).
    
    Xiaomin Chen is with the Department of Computer Science, University of Reading, UK (e-mail: xiaomin.chen@reading.ac.uk)
    }
}

\markboth{Submitted for Review}%
{Shell \MakeLowercase{\textit{et al.}}: Bare Demo of IEEEtran.cls for IEEE Journals}
%

\maketitle

\begin{abstract}
The transition towards sixth-generation (6G) wireless networks necessitates autonomous orchestration mechanisms capable of translating high-level operational intents into executable network configurations. Existing approaches to Intent-Based Networking (IBN) rely upon either rule-based systems that struggle with linguistic variation or end-to-end neural models that lack interpretability and fail to enforce operational constraints. This paper presents a hierarchical multi-agent framework where Large Language Model (LLM) based agents autonomously decompose natural language intents, consult domain-specific specialists, and synthesise technically feasible network slice configurations through iterative reasoning-acting (ReAct) cycles. The proposed architecture employs an orchestrator agent coordinating two specialist agents, namely Radio Access Network (RAN) and Core Network agents, via ReAct-style reasoning, grounded in structured network state representations. Experimental evaluation across diverse benchmark scenarios demonstrates that the proposed system outperforms monolithic agents, rule-based systems, and direct LLM prompting, with architectural principles applicable to Open RAN (O-RAN) deployments.
The results also demonstrate that whilst contemporary LLMs possess substantial general telecommunications knowledge, structured prompt engineering encoding domain-specific decision thresholds and operational policies is essential for reliable network automation performance.
\end{abstract}


\IEEEpeerreviewmaketitle

\section{Introduction}

\IEEEPARstart{T}{he} ongoing transition towards sixth-generation (6G) wireless networks necessitates fundamental advances in autonomous orchestration capabilities. As articulated by Wang \textit{et al.} \cite{wang2023road}, 6G systems envision ubiquitous intelligence and zero-touch network management that transcend the operational paradigms of prior generations. Central to this vision is Intent-Based Networking (IBN), formally defined as a system that translates and validates operational intents into network configurations whilst continuously aligning network behaviour with stated objectives \cite{rfc9315}. However, existing IBN approaches rely upon either structured policy languages that impose accessibility barriers for non-technical operators, or rule-based translation systems that struggle with linguistic variation and fail to generalise beyond explicitly programmed patterns \cite{leivadeas2022survey,pang2020survey}. This semantic gap between natural language intent and technical implementation represents a critical bottleneck for autonomous network management at scale.

Large Language Models (LLMs) offer promising capabilities for bridging this gap through their natural language understanding and generation abilities. Recent surveys have documented LLM applications across telecommunications domains \cite{zhou2024large}, including network configuration generation \cite{wang2024netconfeval}, fault diagnosis, and telecom-specific language model development \cite{zou2025telecomgpt}. However, LLMs exhibit fundamental limitations that constrain their direct applicability to autonomous network orchestration. As demonstrated by Wu \textit{et al.} \cite{wu2024reasoning}, contemporary LLMs excel at pattern recitation rather than genuine reasoning, with performance degrading substantially on counterfactual tasks. More critically, LLMs function as text generation systems lacking autonomous decision-making capabilities. Whilst an LLM can generate syntactically correct configurations, it cannot independently verify feasibility against current network state, coordinate multi-stage provisioning workflows, or adapt dynamically to resource constraints.

Agentic AI systems address above limitations by positioning LLMs as reasoning engines within broader architectures that incorporate planning, tool use, and environmental interaction \cite{wang2024survey}. These systems have demonstrated effectiveness across domains including residential energy management \cite{makroum2025agentic}, robotic control \cite{salimpour2025towards}, and software engineering \cite{roychoudhury2025agentic}. In telecommunications, Jiang \textit{et al.} \cite{jiang2024large} proposed multi-agent frameworks for 6G parameter optimisation, demonstrating cooperative agent coordination through natural language. However, existing work has focused primarily on parameter optimisation rather than autonomous multi-step orchestration workflows requiring iterative reasoning and constraint satisfaction across multiple network domains.

This paper presents a hierarchical agentic AI framework for IBN in 6G environments, where multiple LLM agents collaboratively translate natural language operational intents into executable network slice configurations. The architecture employs an orchestrator agent coordinating two domain-specific specialists, namely Radio Access Network (RAN) and Core Network agents, through ReAct-style reasoning \cite{yao2023react}, with all agents grounded in structured network state representations. The system operates without predefined workflow templates, relying on carefully engineered system prompts to encode domain expertise and operational policies. Experimental evaluation demonstrates that this architecture outperforms monolithic agents, rule-based systems, and direct LLM prompting, whilst revealing that prompt-encoded policy language induces systematic biases in agent behaviour with consequential downstream effects.

The principal contributions of this work are threefold, as outlined below:
\begin{enumerate}
\item We establish a hierarchical multi-agent architecture for IBN orchestration that decomposes the semantic gap problem into tractable sub-problems through domain specialisation and iterative reasoning. Unlike monolithic LLM approaches that combine intent interpretation with resource allocation, the proposed architecture separates semantic reasoning (orchestrator), radio resource policy application (RAN specialist), and latency-aware placement (Core specialist) into distinct agents with formally defined interaction protocols. This decomposition enables independent policy refinement within each domain without requiring end-to-end retraining, a property absent from existing neural IBN approaches.
\item We introduce a hybrid evaluation framework that jointly measures semantic accuracy, quantifying alignment with expert-defined golden standard configurations, and engineering utility, assessing quantitative technical optimality through a weighted multi-criteria function. This dual-metric methodology addresses a methodological gap in agentic system evaluation, where existing approaches rely solely on correctness metrics that fail to capture the engineering quality of generated configurations. The utility function further serves as an intent-sensitive scoring mechanism with dynamically weighted sub-scores conditioned on use case category.
\item We provide empirical evidence demonstrating that prompt-encoded policy language induces systematic and predictable biases in agent behaviour, where seemingly minor lexical variations produce measurable, compounding downstream effects across deployment scenarios. These findings validate prompt engineering as a critical architectural design parameter requiring systematic validation protocols.
\end{enumerate}

The remainder of this paper is organised as follows. Section~\ref{sec:relatedworks} surveys prior work across IBN, machine learning for network optimisation, LLM applications in telecommunications, and agentic AI systems. Section~\ref{sec:model} formalises the system model and problem formulation. Section~\ref{sec:architecture} details the hierarchical agent architecture and implementation. Section~\ref{sec:experiments} presents experimental validation and comparative analysis. Section~\ref{sec:discussion} discusses deployment considerations and prompt engineering challenges. Section~\ref{sec:conclusion} concludes with directions for future research.

\section{Related Works}
\label{sec:relatedworks}
The convergence of IBN paradigms, machine learning advances, and LLM capabilities has created new opportunities for autonomous network orchestration. This section surveys prior work across four interconnected research domains, concluding with identification of the research gap that motivates our hierarchical multi-agent architecture.

\subsection{Intent-Based Networking and Network Automation}
IBN emerged as a paradigm shift towards policy-driven network management, where high-level business objectives are automatically translated into device-specific configurations. Clemm \textit{et al.} \cite{rfc9315} formalised the foundational concepts of IBN, defining intent as a declarative statement of operational goals that abstracts away implementation details. The foundational infrastructure principles were established through Software-Defined Networking (SDN), which introduced programmable control-data plane separation \cite{kreutz2014software}. Leivadeas and Falkner \cite{leivadeas2022survey} conducted the definitive comprehensive survey, establishing systematic taxonomy for intent expression, translation, resolution, activation, and assurance. Their work extensively discusses natural language processing-based intent expression and translation of abstract intents into network configurations, identifying the semantic gap between business-level intents and technical policy representations as a principal limitation.

Pang \textit{et al.} \cite{pang2020survey} established foundational intent-driven network architecture emphasising decoupled control logic for automated intent handling, whilst Benzaid and Taleb \cite{benzaid2020ai} positioned AI-driven automation as the key enabler for zero-touch service management framework. However, these approaches inherited fundamental limitations of requiring structured policy languages or explicit rule specifications rather than natural language interfaces, exhibiting brittleness when encountering linguistic variations not explicitly encoded in their rule sets.

The emergence of network slicing in 5G and beyond introduced additional orchestration complexity. Foukas \textit{et al.} \cite{foukas2017network} established the seminal end-to-end slicing framework across radio access network, transport, and core domains, identifying orchestration complexity as the primary research challenge. Afolabi \textit{et al.} \cite{afolabi2018network} provided comprehensive coverage of SDN-based slicing mechanisms, detailing management and orchestration architectures and network slice templates that directly relate to slice configuration output formats. Coronado \textit{et al.} \cite{coronado2022zero} surveyed zero-touch management solutions, explaining why human-in-the-loop management fails at 6G scale and justifying the necessity for autonomous intent translation mechanisms.

Recent O-RAN-based orchestration frameworks have demonstrated intent-driven automation capabilities. D'Oro \textit{et al.} \cite{d2023orchestran} introduced OrchestRAN, which translates high-level operator control objectives into optimised near-real-time application (xApp) and non-real-time application (rApp) deployments, formulating intent realisation as an NP-hard optimisation problem and achieving sub-second computation times for large-scale networks. Similarly, Bonati \textit{et al.} \cite{bonati2023neutran} proposed NeutRAN, implementing zero-touch automation pipelines that provision complete network slices in approximately 10 seconds based on operator requests. However, both systems operate within structured intent specification frameworks requiring predefined parameters or templates. Whilst these approaches excel at resource optimisation given explicit numerical constraints, they lack the semantic reasoning capabilities necessary to process unstructured natural language expressions containing implicit requirements. Our work addresses this limitation by positioning LLMs as the semantic interpretation layer, enabling genuine natural language interfaces whilst maintaining automated orchestration principles.

\subsection{Machine Learning for Network Optimisation}
The application of machine learning techniques to network optimisation has evolved from classical analytical methods such as convex optimization to contemporary deep learning approaches. Deep reinforcement learning has emerged as the dominant paradigm for dynamic network management. Luong \textit{et al.} \cite{luong2019applications} conducted the foundational survey covering deep reinforcement learning applications across network layers including dynamic spectrum access, resource allocation, and traffic routing. Zhang \textit{et al.} \cite{zhang2019deep} provided encyclopedic coverage of deep learning across the protocol stack, from physical layer modulation to resource allocation. These approaches demonstrated performance improvements over manually designed algorithms but faced fundamental limitations for IBN applications.

Hua \textit{et al.} \cite{hua2019gan} proposed GAN-powered distributional reinforcement learning for demand-aware network slicing, Yan \textit{et al.} \cite{yan2019intelligent} applied deep reinforcement learning to intelligent resource scheduling for RAN slicing, whilst Eisen and Ribeiro \cite{eisen2020optimal} introduced random edge graph neural networks exploiting network topology for scalable resource allocation. However, these methods require numerical state representations that preclude direct processing of natural language intents. Training necessitates extensive simulation or real-world interaction, which is infeasible for safety-critical telecommunications infrastructure. Perhaps most critically for operational deployment, learned policies function as black-box decision systems lacking the interpretability required in regulated environments. Our work addresses these limitations by leveraging LLMs' natural language understanding capabilities whilst maintaining grounded reasoning through structured network state representations.

\subsection{Large Language Models in Communications}
The rapid advancement of foundation models has created new opportunities for natural language interfaces to complex technical systems. Zhou \textit{et al.} \cite{zhou2024large} conducted the definitive survey documenting LLM applications in telecommunications, establishing principles for LLM deployment including architecture design, pre-training strategies, and evaluation methodologies. Their comprehensive taxonomy covers generation, classification, optimisation, and prediction tasks, providing foundational context for LLM-enabled telecom applications.

Recent work has explored LLM adaptation specifically for networking domains. Wu \textit{et al.} \cite{wu2024netllm} introduced NetLLM, demonstrating the first framework for adapting LLMs to networking tasks with multimodal encoders and efficient fine-tuning, achieving 10 to 40 percent improvements over state-of-the-art methods on diverse tasks. Zou \textit{et al.} \cite{zou2025telecomgpt} established systematic pipelines for continual pre-training and instruction tuning to build telecom-specific models, whilst Wang \textit{et al.} \cite{wang2024netconfeval} provided systematic evaluation of LLM capabilities for network configuration generation, identifying limitations and proposing design principles for LLM-based configuration systems.

Bariah \textit{et al.} \cite{bariah2024large} articulated the vision of large generative models as enablers of autonomous wireless networks where multi-modal models eliminate task-specific AI requirements. Maatouk \textit{et al.} \cite{maatouk2024large} provided practical industry-focused assessment of LLM readiness for anomaly resolution and technical specification comprehension. Mekrache \textit{et al.} \cite{mekrache2024intent} proposed LLM-centric Intent Life-Cycle management architecture addressing complete intent lifecycle including decomposition, translation, negotiation, activation, and assurance. However, these applications primarily utilised LLMs as auxiliary tools for generating candidate solutions or single-shot prompting for isolated tasks, rather than autonomous orchestration engines employing iterative multi-agent reasoning protocols.

\subsection{Agentic AI and Multi-Agent Systems}
Agentic AI systems extend beyond passive text generation to encompass autonomous goal-directed behaviour, tool use, and environmental interaction. The ReAct paradigm introduced by Yao \textit{et al.} \cite{yao2023react} demonstrated that interleaving reasoning traces with action execution enables iterative task refinement whilst producing interpretable decision traces. Wei \textit{et al.} \cite{wei2022chain} established that intermediate reasoning steps enable complex multi-step reasoning through Chain-of-Thought prompting, providing foundational principles for orchestrator-specialist iterative refinement. Schick \textit{et al.} \cite{schick2023toolformer} introduced Toolformer, demonstrating LLMs' capacity for autonomous tool selection and invocation without sacrificing core language capabilities.

Wang \textit{et al.} \cite{wang2024survey} provided unified architectural framework for LLM-based autonomous agents encompassing profiling, memory, planning, and action modules directly applicable to describing orchestrator and specialist agent architectures. Li \textit{et al.} \cite{li2023camel} introduced role-playing and inception prompting mechanisms for autonomous cooperation between LLM agents, discussing challenges including looping behaviours and role drift in multi-agent systems. Xi \textit{et al.} \cite{xi2025rise} surveyed single-agent, multi-agent, and human-agent cooperation scenarios, contextualising hierarchical agent architectures applicable to distributed network management.

However, a critical challenge identified across LLM applications is reliability and systematic bias mitigation. Si \textit{et al.} \cite{si2022prompting} decomposed reliability into generalizability, social biases, calibration, and factuality, establishing empirically validated prompting methods for mission-critical applications. This work directly addresses prompt engineering challenges identified in our experimental findings, where seemingly minor linguistic variations in system prompts induce measurable behavioural differences that compound across deployment scenarios.

\subsection{Research Gap and Positioning}
The literature reveals a persistent gap between IBN requirements and current capabilities. Traditional IBN systems require structured policy languages imposing accessibility barriers \cite{leivadeas2022survey,pang2020survey}. Machine learning approaches for network optimisation demonstrate strong performance on numerical tasks but cannot process natural language intents and lack interpretability \cite{luong2019applications,zhang2019deep}. Recent LLM applications in telecommunications have focused primarily on single-shot prompting \cite{zhou2024large,wu2024netllm} or auxiliary configuration assistance \cite{wang2024netconfeval,zou2025telecomgpt} rather than autonomous multi-step orchestration workflows.

Existing agentic AI frameworks \cite{yao2023react,wang2024survey,li2023camel}, whilst demonstrating effectiveness in domains such as energy management and software development, have not been systematically applied to IBN orchestration. The telecommunications domain presents unique challenges including safety-critical operation requirements, strict latency constraints, necessity for continuous adaptation to dynamic network states, and regulatory demands for decision auditability. The 6G vision articulated by Saad \textit{et al.} \cite{saad2019vision} and Wang \textit{et al.} \cite{wang2023road} emphasises autonomous network management and AI-driven orchestration as essential requirements, yet practical implementations employing LLM-based agentic reasoning remain largely unexplored. This work addresses this gap by presenting a hierarchical multi-agent architecture specifically designed for IBN in 6G environments, where domain-specific specialist agents collaboratively translate natural language operational intents into executable network slice configurations through iterative reasoning protocols.

\section{System Model and Problem Formulation}
\label{sec:model}

This section formalises the IBN orchestration problem for 6G environments. We begin by characterising the standardised use case categories that define Quality of Service (QoS) requirements, then establish the formal network system model, introduce the evaluation framework, and finally present the mathematical problem formulation.

\subsection{6G Use Case Categories and QoS Requirements}
\label{sec:usecase}

The International Telecommunication Union (ITU) has defined foundational service categories for next-generation wireless systems that establish distinct QoS profiles. The ITU-R framework for IMT-2030 \cite{itu2023framework} expands 5G's three usage scenarios, namely Ultra-Reliable Low-Latency Communication (URLLC), enhanced Mobile Broadband (eMBB), and massive Machine-Type Communication (mMTC), into six distinct 6G categories incorporating immersive communication, hyper-reliable low-latency communication, and integrated AI and communication \cite{liu2025itu}. This expansion, combined with network densification, heterogeneous spectrum management spanning sub-6 GHz through terahertz bands, and the proliferation of AI-native network functions, substantially amplifies orchestration complexity beyond what 5G systems demanded. We adopt the foundational three-category framework for evaluation, as the fundamental QoS trade-offs that distinguish latency-critical, capacity-critical, and connectivity-critical applications underpin all six evolved categories. The URLLC, eMBB, mMTC terminology further maintains widespread recognition in telecommunications research and operational practice, facilitating clear communication of evaluation scope.

\textbf{Ultra-Reliable Low-Latency Communication (URLLC)} encompasses safety-critical and time-sensitive applications requiring end-to-end latency below 10 milliseconds and reliability exceeding 99.999 percent \cite{durisi2016toward,3gpp2018study}. Representative use cases include industrial automation, autonomous vehicle coordination, and remote surgical procedures \cite{bennis2018ultrareliable}. URLLC applications mandate edge-tier compute placement to minimise propagation delay and require guaranteed resource allocation to prevent packet loss during congestion.

\textbf{Enhanced Mobile Broadband (eMBB)} addresses high-capacity applications requiring substantial throughput whilst tolerating moderate latency (20 to 50 milliseconds) \cite{shafi20175g}. Characteristic use cases include ultra-high-definition video streaming, immersive extended reality, and fixed wireless access \cite{wang2023road}. eMBB applications can leverage regional-tier data centres where network buffering masks propagation latency, prioritising spectrum efficiency and aggregate capacity over minimal delay.

\textbf{Massive Machine-Type Communication (mMTC)} supports extremely high device densities with low per-device data rates and high latency tolerance (seconds to minutes) \cite{3gpp2018study}. Typical deployments include smart metering infrastructure, agricultural sensor networks, and environmental monitoring systems. mMTC applications favour wide-area coverage and energy-efficient transmission, often utilising low-band spectrum for superior propagation characteristics.

These categorical distinctions inform the design of our evaluation framework (Section~\ref{sec:evaluation}) and enable context-dependent weighting of quality metrics.

\subsection{Network System Model}
\label{sec:sysmodel}

We consider a 6G network environment characterised by a state space $\mathcal{S}$ comprising a set of RAN sectors $\mathcal{R}$ and a set of Core Network compute nodes $\mathcal{C}$. Each RAN sector $r \in \mathcal{R}$ is defined by a tuple $(L_r, U_r, B_r)$, where $L_r \in [0, 100]$ represents the current load percentage, $U_r \in \mathbb{Z}^+$ denotes the number of active users, and $B_r = \{b_{\text{mmWave}}, b_{\text{mid}}, b_{\text{low}}\}$ specifies available spectrum capacity (in MHz) across millimetre-wave, mid-band, and low-band frequency ranges.

\begin{figure}[!t]
    \centering
    \includegraphics[width=\columnwidth]{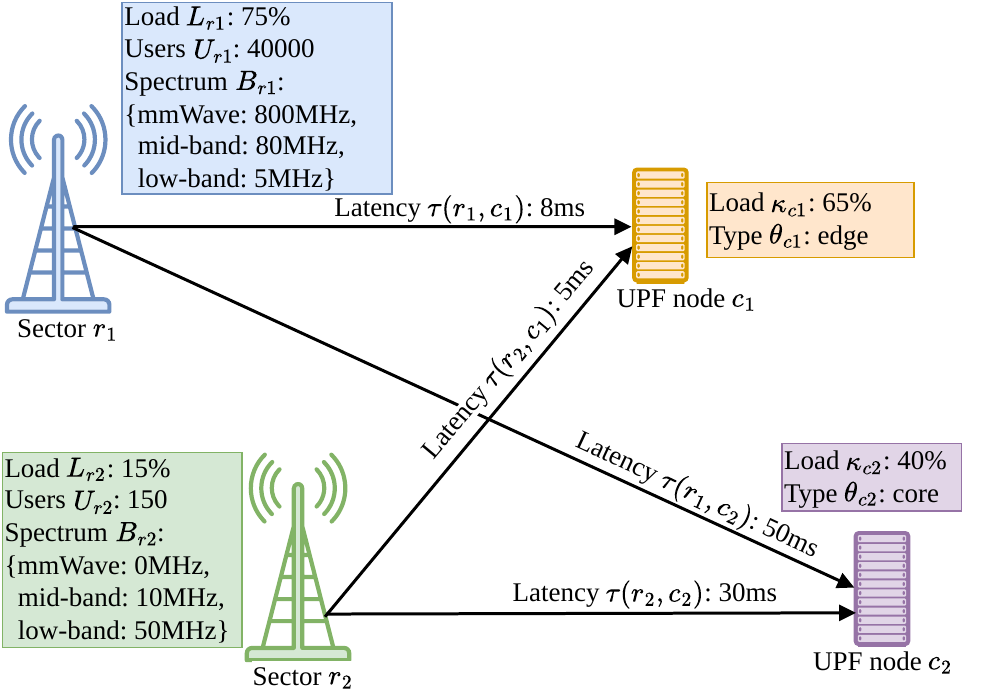}
    \caption{Network system model illustration showing RAN sectors characterised by load $L_r$, active users $U_r$, and spectrum availability $B_r$, alongside core nodes characterised by compute load $\kappa_c$ and tier $\theta_c$. The latency matrix $\mathbf{T}$ captures end-to-end latency $\tau(r,c)$ between each sector-node pair.}
    \label{fig:system_model}
\end{figure}

Similarly, each core node $c \in \mathcal{C}$ is characterised by $(\kappa_c, \theta_c)$, where $\kappa_c \in [0, 100]$ represents computational load as a percentage of available capacity, and $\theta_c \in \{\text{edge}, \text{core}\}$ indicates the node's hierarchical tier in the network topology. The end-to-end latency between RAN sector $r$ and core node $c$ is denoted by $\tau(r,c) \in \mathbb{R}^+$ (measured in milliseconds), capturing geographic distance and topological routing effects. The complete set of latencies forms a latency matrix $\mathbf{T} = [\tau(r,c)]_{r \in \mathcal{R}, c \in \mathcal{C}}$ that encodes sector-to-node connectivity characteristics. Fig.~\ref{fig:system_model} illustrates this network system model with two example RAN sectors and two User Plane Function (UPF) nodes, showing how the latency matrix encodes sector-to-node connectivity.

The objective of the IBN system is to process a natural language intent $I$ provided by a human operator and synthesise an optimal provisioning configuration $\mathcal{A}^*$. This configuration is a structured tuple:
\begin{equation}
\mathcal{A}^* = (r^*, b^*, c^*)
\end{equation}
where $r^* \in \mathcal{R}$ is the selected RAN sector, $b^* \in \{\text{mmWave}, \text{mid-band}, \text{low-band}\}$ is the allocated spectrum band, and $c^* \in \mathcal{C}$ is the designated UPF node. The configuration $\mathcal{A}^*$ needs to satisfy the QoS requirements implicit in $I$ whilst adhering to the resource constraints encoded in $\mathcal{S}$.

\subsection{Evaluation Framework}
\label{sec:evaluation}

Evaluating agentic systems in qualitative domains such as IBN presents methodological challenges due to the absence of a single objectively correct configuration for a given intent. To address this, we introduce a hybrid evaluation framework consisting of two complementary metrics: Semantic Accuracy (i.e., qualitative alignment with expert expectations) and Engineering Utility (i.e., quantitative assessment of technical optimality).

\subsubsection{Semantic Accuracy, i.e., Golden Standard}

We define a benchmark set of representative test scenarios spanning diverse 6G use cases. Each scenario consists of a natural language intent and a corresponding expert-defined ideal configuration. This ideal configuration, termed the ``Golden Standard'', is specified by human domain experts with knowledge of RAN resource allocation and core network topology.

System performance is evaluated by comparing the agent-generated configuration $\mathcal{A}^*$ against the Golden Standard. We define a three-level semantic accuracy score:
\begin{equation}
A(\mathcal{A}^*, \mathcal{A}_{\text{golden}}) = \begin{cases}
1.0 & \text{if } \mathcal{A}^* = \mathcal{A}_{\text{golden}} \\
0.5 & \text{if } \mathcal{A}^* \approx \mathcal{A}_{\text{golden}} \\
0.0 & \text{otherwise}
\end{cases}
\end{equation}

A Perfect Match (1.0) indicates that the system's output exactly matches the expert configuration across all three dimensions: sector, spectrum, and UPF node. A Partial Match (0.5) is assigned when the configuration matches on two of three dimensions. No Match (0.0) indicates fundamental misalignment with expert expectations.

\subsubsection{Engineering Utility Function}

To quantify the technical quality of provisioned slices beyond binary correctness, we define a numerical Utility Function $U$ that scores configurations against key network performance indicators. The utility function is formulated as a weighted sum of three normalised sub-scores:
\begin{equation}
U(\mathcal{A}, I, \mathcal{S}) = w_l \cdot S_{\text{latency}} + w_r \cdot S_{\text{resource}} + w_c \cdot S_{\text{congestion}}
\label{eq:uscore}
\end{equation}
where $\mathcal{A}$ represents any candidate configuration being evaluated, and $w_l$, $w_r$, and $w_c$ are non-negative weights satisfying $w_l + w_r + w_c = 1$. The utility function $U$ accepts three arguments: the configuration $\mathcal{A}$ to be evaluated, the natural language intent $I$ that determines context-dependent weighting, and the current network state $\mathcal{S}$ that provides resource availability information.

The latency score $S_{\text{latency}} \in [0, 1]$ evaluates the appropriateness of UPF node selection relative to the intent's latency requirements. Given a configuration $\mathcal{A}^* = (r^*, b^*, c^*)$, we assess the end-to-end latency $\tau(r^*,c^*)$ from the selected RAN sector $r$ to the chosen core node $c$:

\begin{equation}
S_{\text{latency}} = \begin{cases}
1.0 & \text{if } \tau(r^*, c^*) < 10 \text{ ms } \\
0.5 & \text{if } 10 \text{ ms } \leq \tau(r^*, c^*) < 30 \text{ ms } \\
0.0 & \text{if } \tau(r^*, c^*) \geq 30 \text{ ms )}
\end{cases}
\end{equation}

The resource efficiency score $S_{\text{resource}} \in [0, 1]$ penalises spectrum allocation decisions that waste scarce resources. We first classify the bandwidth requirement category $\beta(I)$ from intent $I$ based on use case keywords. For a given configuration $\mathcal{A}^* = (r^*, b^*, c^*)$, the resource efficiency score evaluates whether the selected spectrum band $b^*$ appropriately matches the inferred requirement category $\beta(I)$:

\begin{equation}
S_{\text{resource}} = \begin{cases}
1.0 & \text{if } b^* = \text{mmWave} \text{ and } \beta(I) = \text{high} \\
1.0 & \text{if } b^* = \text{mid-band} \text{ and } \beta(I) \neq \text{high} \\
1.0 & \text{if } b^* = \text{low-band} \text{ and } \beta(I) = \text{low} \\
0.5 & \text{if } b^* = \text{mmWave} \text{ and } \beta(I) = \text{medium} \\
0.5 & \text{if } b^* = \text{mid-band} \text{ and } \beta(I) = \text{high} \\
0.5 & \text{if } b^* = \text{low-band} \text{ and } \beta(I) \neq \text{low} \\
0.0 & \text{if } b^* = \text{mmWave} \text{ and } \beta(I) = \text{low}
\end{cases}
\end{equation}

The congestion score $S_{\text{congestion}} \in [0, 1]$ evaluates the load level of the selected RAN sector. For configuration $\mathcal{A}^* = (r^*, b^*, c^*)$:

\begin{equation}
S_{\text{congestion}} = 1 - \frac{L_{r^*}}{100}
\end{equation}
where $L_{r^*}$ is the current load percentage of the selected sector.

The weights $w_l$, $w_r$, and $w_c$ are dynamically assigned based on the semantic content of the intent $I$ and the use case category identified in Section~\ref{sec:usecase}. For URLLC-oriented intents (containing keywords such as ``ultra-low latency'' or ``real-time''), we set $w_l = 0.8$, $w_r = 0.1$, $w_c = 0.1$ to heavily prioritise latency compliance. For capacity-oriented eMBB intents, we set $w_c = 0.6$, $w_r = 0.3$, $w_l = 0.1$ to emphasise congestion avoidance and resource efficiency. For coverage-oriented mMTC intents, we set $w_r = 0.6$, $w_c = 0.3$, $w_l = 0.1$ to prioritise appropriate spectrum selection.

\subsubsection{Hybrid Evaluation Protocol}

For each test scenario, we execute the system and record both the Semantic Accuracy score $A$ and the Engineering Utility score $U$. A successful deployment satisfies two conditions: semantic alignment with expert expectations ($A \geq 0.5$) and engineering feasibility ($U \geq 0.7$). This dual-metric approach enables us to verify that the agent not only understood the operator's intent but also implemented it optimally given the prevailing network conditions.

\subsection{Problem Formulation}

Given the network system model discussed in Section~\ref{sec:sysmodel} and evaluation framework discussed in Section~\ref{sec:evaluation}, we now formally state the IBN orchestration problem. The system seeks an optimal configuration that maximises the utility function whilst satisfying operational constraints:

\begin{subequations}
\begin{equation}
\label{eq:optimisation}
\mathcal{A}^* = \arg\max_{\mathcal{A}} \, U(\mathcal{A}, I, \mathcal{S})
\end{equation}
\begin{align}
\text{s.t.} \quad
L_{r^*} &\leq L_{\text{max}} \\
b^* &\in B_{r^*} \\
\kappa_{c^*} &\leq \kappa_{\text{max}} \\
\tau(r^*, c^*) &\leq \tau_{\text{req}}(I)
\end{align}
\end{subequations}
where $U(\mathcal{A}, I, \mathcal{S})$ is the utility function defined in Equation~(\ref{eq:uscore}), $L_{\text{max}}$ is the sector's load threshold, $B_{r^*}$ denotes available spectrum bands, $\kappa_{\text{max}}$ is the computation capacity of the UPF node, and $\tau_{\text{req}}(I)$ denotes the latency threshold extracted from the operator's intent.

Unlike classical optimisation problems where objectives are expressed as mathematical functions, the intent $I$ in IBN systems is unstructured natural language (e.g., ``Configure network for a sports tournament requiring ultra-low latency and high user density''). This necessitates a semantic reasoning layer capable of mapping linguistic abstractions (such as ``ultra-low latency'' or ``high density'') to concrete technical constraints and subsequently resolving them against the current network state $\mathcal{S}$.

The challenge lies not in solving a constrained optimisation problem with known parameters, but rather in the interpretation and extraction of those parameters from natural language. Traditional mathematical optimisation techniques assume that decision variables, objective functions, and constraints are explicitly specified. In contrast, IBN requires semantic understanding to identify which constraints are relevant (e.g., latency versus capacity), extract implicit numerical thresholds (e.g., mapping ``ultra-low latency'' to $\tau < 10$ ms), and infer use case categories that determine utility function weights.

This semantic gap motivates the agentic AI architecture presented in Section~\ref{sec:architecture}, in which LLMs serve as the reasoning layer that bridges natural language intent and formal technical specifications. Rather than solving the optimisation problem in Equation~(\ref{eq:optimisation}) directly, the agents approximate constraint-satisfying solutions through prompt-encoded expert decision rules that map natural language semantics to technical decisions. The utility function $U(\mathcal{A}, I, \mathcal{S})$ is then applied after configuration generation to quantify how closely each output approaches the theoretical optimum, enabling systematic comparison across architectural configurations and ablation conditions.

\section{Agentic AI Architecture and Implementation}
\label{sec:architecture}

This section details the hierarchical multi-agent architecture and its practical implementation. We describe the agent organisation, coordination protocol, prompt engineering strategy, network state representation, and iterative reasoning mechanism that collectively enable autonomous intent translation.

The system accepts as input a natural language intent string $I$ provided by a human operator, such as ``\textit{Configure network slice for automated robotic assembly line at industrial\_park\_a requiring ultra-low latency (<5ms) for real-time control and high reliability for safety-critical operations.}'' The system produces as output a structured provisioning configuration $\mathcal{A}^* = (r^*, b^*, c^*)$ specifying the selected RAN sector, spectrum band, and UPF node, accompanied by a complete reasoning trace documenting the decision rationale at each stage. The agents possess no prior training data, fine-tuning, or external knowledge bases. Their domain knowledge derives entirely from two sources: (1)~expert-encoded policy rules within system prompts (detailed in Section~\ref{sec:prompts}), and (2)~structured network state $\mathcal{S}$ injected at inference time (detailed in Section~\ref{sec:state}). The orchestrator agent has access to two actions: \texttt{call\_agent} for consulting a specialist with a natural language query and the current network state, and \texttt{provision\_slice} for executing the final configuration. Specialist agents possess no direct tool access, functioning as constrained reasoning engines that receive context from the orchestrator and return natural language recommendations.

\subsection{Hierarchical Multi-Agent Architecture}

\begin{figure}[!b]
    \centering
	\includegraphics[width=\columnwidth]{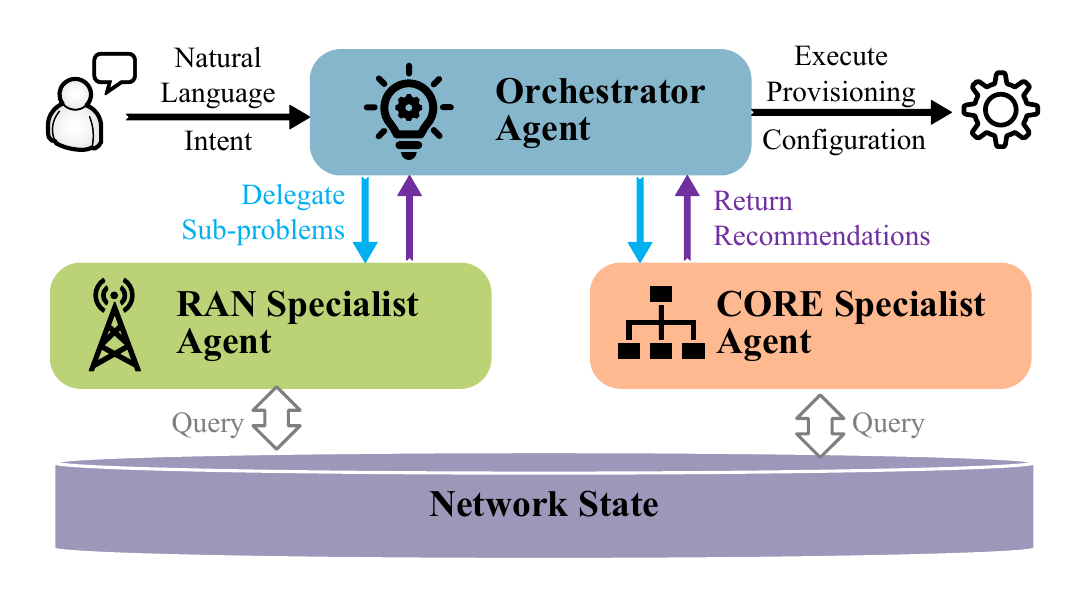}
    \caption{Hierarchical multi-agent architecture for IBN orchestration.}
    \label{fig:architecture}
\end{figure}

To address the semantic reasoning and constraint satisfaction challenges identified in Section~\ref{sec:model}, we propose a hierarchical agentic AI architecture inspired by the ReAct paradigm \cite{yao2023react}. The system comprises three LLM-based agents organised in a two-tier structure: a central orchestrator agent and two domain-specific specialist agents, as shown in Fig.~\ref{fig:architecture}.

\subsubsection{Orchestrator Agent}
The orchestrator serves as the central decision-making entity and the unified interface for human operators. Its primary responsibilities include intent parsing (i.e., decomposing the high-level intent $I$ into domain-specific sub-problems), delegation to appropriate specialists, synthesis of recommendations, and execution of final configurations. The orchestrator operates through iterative Thought-Action-Observation cycles, generating reasoning traces, executing tool calls, and refining its plan based on feedback.

\subsubsection{Specialist Agents}
To mitigate context window limitations and reasoning bottlenecks inherent in monolithic LLM systems, we offload domain-specific reasoning to specialised agents. These agents do not possess direct tool access; instead, they function as constrained reasoning engines that receive environmental context from the orchestrator.

\textit{The RAN Specialist Agent} embodies expertise in radio resource management, with policies for spectrum allocation based on user density and propagation requirements. The agent receives a structured representation of the relevant RAN sector's state (including current load, active user count, and available spectrum) and returns a justified recommendation in natural language.

\textit{The Core Specialist Agent} specialises in network topology and UPF placement, with guidelines for latency optimisation and compute resource allocation. Upon receiving a query from the orchestrator, the Core specialist evaluates candidate UPF nodes against latency and capacity constraints, returning a placement recommendation with justification.

This modular design facilitates independent refinement of specialist knowledge without necessitating retraining of the central orchestrator.

\subsection{Sequential Coordination Protocol}
\label{sec:coordination}

\begin{figure}[!tb]
    \centering
	\includegraphics[width=\columnwidth]{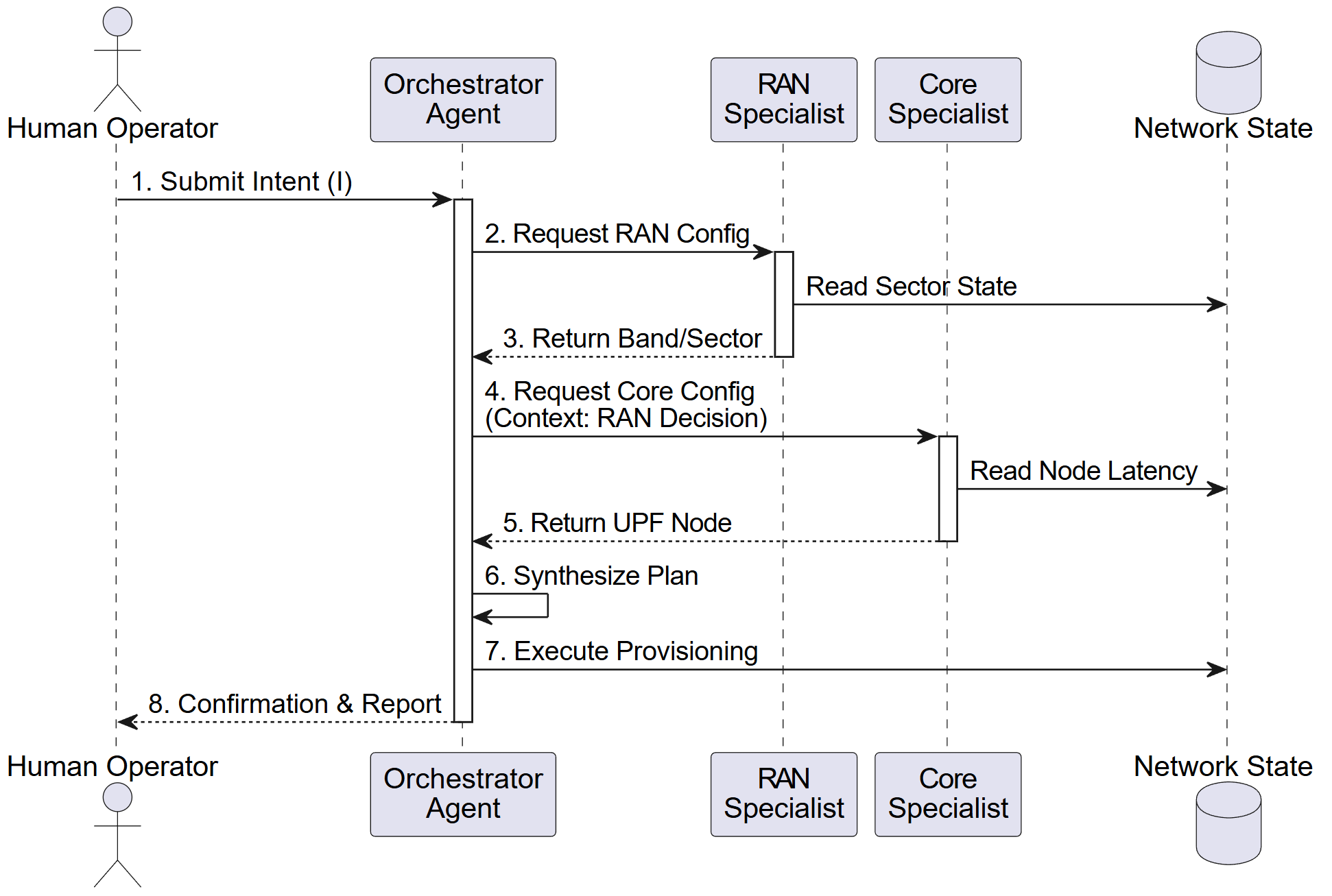}
    \caption{Sequential coordination protocol for multi-domain network slice provisioning.}
    \label{fig:protocol}
\end{figure}

As shown in Fig.~\ref{fig:protocol}, the system employs a sequential consultation protocol to resolve multi-domain constraints systematically. During initialisation, the orchestrator receives intent $I$ from the human operator and parses it to identify key requirements (e.g., latency thresholds, user density, reliability constraints). The orchestrator then consults the RAN specialist to establish physical layer feasibility. The query specifies the target sector extracted from the intent and asks whether the sector can accommodate the implied traffic load and QoS requirements. The RAN specialist examines sector load $L_r$ and available spectrum $B_r$, returning a spectrum band recommendation $b^*$ or a warning if the sector is congested.

Conditioned on the RAN recommendation, the orchestrator proceeds to consult the Core specialist. The query incorporates the RAN decision to ensure compatibility. The Core specialist evaluates candidate nodes based on their latency profiles to the target sector from the latency matrix $\mathbf{T}$ and compute availability $\kappa_c$, selecting the optimal UPF node $c^*$. The orchestrator then aggregates the recommendations $(r^*, b^*, c^*)$ and performs a final consistency check. This includes verifying that the selected UPF node achieves acceptable latency to the RAN sector and that no resource conflicts exist. Upon validation, the orchestrator invokes the provisioning tool, which applies the configuration to the network controller.

This decoupled, sequential approach offers three principal advantages. First, interpretability is enhanced as each agent's reasoning trace is preserved, enabling operators to audit decision rationale. Second, modularity allows specialist agents to be independently upgraded or replaced without disrupting the orchestrator. Third, constraint propagation is achieved by conditioning Core decisions on RAN outputs, ensuring that inter-domain dependencies are respected. However, this sequential protocol assumes that RAN and Core decisions are largely independent given the intent, an assumption that may not hold for certain advanced slicing scenarios requiring joint optimisation across domains.

\subsection{Agent Prompt Engineering}
\label{sec:prompts}

The core intelligence of our system resides in the prompt engineering of the three LLM agents. We employ an agent-specific prompting strategy, in which each agent is assigned a specific expert role and a set of operational constraints encoded directly within its system prompt.

The orchestrator is initialised with a system prompt that defines its role as a ``6G Intent Coordinator.'' It is explicitly instructed to refrain from making technical decisions directly but instead to delegate domain-specific sub-problems to appropriate specialists. The prompt includes a structured description of available tools (\texttt{call\_agent} for specialist consultation and \texttt{provision\_slice} for configuration execution) and encodes the ReAct protocol through explicit formatting guidelines. The orchestrator's prompt instructs it to generate structured outputs in the following format:

\begin{verbatim}
THOUGHT: [Reasoning about current state]
ACTION: [Tool invocation with parameters]
OBSERVATION: [Result from tool execution]
\end{verbatim}

This structured generation ensures that the orchestrator's decision-making process remains transparent and auditable. The prompt further specifies coordination rules, such as ``ALWAYS consult specialists before provisioning'' and ``Synthesise recommendations from multiple agents before final decision,'' which prevent premature configuration application.

The specialist agents (RAN and Core) are prompted with domain-specific policies that encode expert knowledge. The RAN specialist prompt includes quantitative decision rules. For instance, it specifies: ``For high user density scenarios (exceeding 10k active users), prioritise millimetre-wave spectrum if available, as mid-band cannot handle the capacity.'' The prompt further encodes reliability policies: ``For high-reliability or wide-coverage requirements, favour mid-band spectrum due to its superior propagation characteristics compared to mmWave.'' Additionally, the prompt instructs the agent to issue congestion warnings: ``If a sector's load percentage exceeds 80 percent, explicitly warn the orchestrator that the sector is approaching capacity limits.''

The Core specialist prompt encodes latency-centric policies. It specifies: ``For URLLC intents requiring latency below 10 milliseconds, specify the selection of multi-access edge computing (MEC) nodes. Regional or centralised data centres cannot satisfy such stringent latency constraints.'' Conversely, for standard latency requirements, the prompt instructs the agent to recommend regional data centres to optimise cost-efficiency by avoiding over-utilisation of expensive edge resources. The prompt further includes capacity verification rules: ``Before recommending a UPF node, verify that its current compute load is below 85 percent to ensure sufficient headroom for the new slice.''

Each specialist prompt concludes with an explicit output format, requiring that recommendations be structured as:

\begin{verbatim}
RECOMMENDATION: [Action] at [Location] 
because [Reason].
\end{verbatim}

This enforced structure ensures specialist outputs are consistently parsable by the orchestrator and provide rationale for audit purposes. All prompts are stored as Markdown files external to the codebase, enabling rapid iteration and version control of agent behaviours.

\subsection{Network State Representation}
\label{sec:state}

The agents interact with the network through structured representations of network state $\mathcal{S}$. For experimental validation, we implement a simplified network state model as a structured JSON object. This approach enables controlled evaluation of the agentic system's reasoning capabilities without the complexity of interfacing with production network management platforms. The JSON representation serves as a mock network state that captures the essential characteristics defined in our system model whilst abstracting away implementation details of actual telecommunications infrastructure.

The structured representation contains two primary components. The \texttt{sectors} object maps sector identifiers to their current operational state. Each sector entry includes \texttt{active\_users} (the count of connected devices), \texttt{load\_percentage} (current utilisation as a percentage of maximum capacity), and \texttt{spectrum\_available\_mhz} (a nested object specifying available capacity in megahertz for each frequency band: \texttt{mmWave}, \texttt{mid\_band}, and \texttt{low\_band}). The \texttt{nodes} object similarly maps node identifiers to their operational characteristics. Each node entry includes \texttt{type} (classification as \texttt{edge}, \texttt{regional}, or \texttt{core}), \texttt{latency\_to\_ran\_ms} (end-to-end latency from the RAN edge in milliseconds), and \texttt{compute\_load\_percent} (current computational utilisation).

During execution, when a specialist agent is invoked, the system loads the agent's system prompt and appends the serialised network state to the query, formatted as:

\begin{verbatim}
ORCHESTRATOR REQUEST:
[Query from Orchestrator]

CURRENT NETWORK STATE:
[JSON representation of network state]

Provide your expert recommendation based 
on this state.
\end{verbatim}

This injection mechanism ensures that the specialist's reasoning is grounded in verified network conditions. The LLM receives both the task description and the complete environmental state within a single context window, enabling it to generate recommendations that respect physical constraints without requiring iterative tool calls. This context grounding prevents hallucinations regarding resource availability.

\subsection{ReAct Loop Implementation}

The orchestrator's iterative reasoning loop is encapsulated within a dedicated coordination module. Upon receiving a user intent, the orchestrator initialises a conversation history with two messages: a system message containing the orchestrator's prompt and a user message containing the intent. The system then enters a loop (with a maximum iteration limit of 10 to prevent infinite loops) where it repeatedly calls the underlying LLM API to generate the next thought or action. The complete reasoning procedure is formalised in Algorithm~\ref{alg:react}.

\begin{algorithm}[!b]
\caption{Orchestrator ReAct Reasoning Loop}
\label{alg:react}
\begin{algorithmic}[1]
\REQUIRE Natural language intent $I$, maximum iterations $K_{\text{max}}=10$
\ENSURE Provisioned configuration $\mathcal{A}^*$ or failure report
\STATE Initialise conversation history $H \leftarrow [\text{SystemPrompt}, I]$
\FOR{$k = 1$ to $K_{\text{max}}$}
    \STATE $r \leftarrow \text{QueryLLM}(H, \text{temperature}=0.0)$
    \STATE $H \leftarrow H \cup \{r\}$
    \STATE $\text{actionType} \leftarrow \text{ParseAction}(r)$
    \IF{$\text{actionType} = \text{FINISH}$}
        \STATE $\mathcal{A}^* \leftarrow \text{ExtractConfiguration}(r)$
        \RETURN $\mathcal{A}^*$
    \ELSIF{$\text{actionType} = \text{CALL\_AGENT}$}
        \STATE $(a, q) \leftarrow \text{ExtractParameters}(r)$
        \STATE $\text{obs} \leftarrow \text{ConsultSpecialist}(a, q, \mathcal{S})$
        \STATE $H \leftarrow H \cup \{\text{``Observation: ''} + \text{obs}\}$
    \ELSIF{$\text{actionType} = \text{PROVISION\_SLICE}$}
        \STATE $\text{config} \leftarrow \text{ExtractParameters}(r)$
        \STATE $\text{ExecuteProvisioning}(\text{config})$
    \ENDIF
\ENDFOR
\RETURN Failure (iteration limit exceeded)
\end{algorithmic}
\end{algorithm}

\begin{table*}[!b]
\caption{RAN Sector Topology Characteristics}
\label{tab:ran_sectors_topology}
\centering
\begin{tabular}{lcccccc}
\toprule
\textbf{Sector ID} & \textbf{Active} & \textbf{Load} & \multicolumn{3}{c}{\textbf{Spectrum Available (MHz)}} & \textbf{Context} \\
& \textbf{Users} & \textbf{(\%)} & \textbf{mmWave} & \textbf{Mid} & \textbf{Low} & \\
\hline
\texttt{stadium\_central} & 45,000 & 88 & 800 & 20 & 5 & High-density urban venue \\
\texttt{city\_plaza} & 1,200 & 35 & 800 & 80 & 30 & Commercial district \\
\texttt{industrial\_park\_a} & 500 & 60 & 400 & 100 & 20 & Factory environment \\
\texttt{suburban\_residential} & 3,000 & 45 & 0 & 60 & 40 & Fixed wireless access \\
\texttt{rural\_highway} & 150 & 15 & 0 & 20 & 50 & Wide-area coverage \\
\bottomrule
\end{tabular}
\begin{flushleft}
\footnotesize
Note: Load percentage represents current utilisation relative to maximum sector capacity. Spectrum availability reflects unallocated bandwidth in each frequency band.
\end{flushleft}
\end{table*}

\begin{table*}[!b]
\centering
\caption{Core Network UPF Node Characteristics}
\label{tab:upf_nodes_topology}
\begin{tabular}{lccccccc}
\toprule
& \multicolumn{5}{c}{\textbf{Latency to RAN Sector (ms)}} & & \\
\cmidrule(lr){2-6}
\textbf{Node ID} & \textbf{stadium\_} & \textbf{city\_} & \textbf{industrial\_} & \textbf{suburban\_} & \textbf{rural\_} & \textbf{Compute} & \textbf{Description} \\
& \textbf{central} & \textbf{plaza} & \textbf{park\_a} & \textbf{residential} & \textbf{highway} & \textbf{Load (\%)} & \\
\midrule
\texttt{mec\_stadium\_1} & 2 & 12 & 15 & 25 & 45 & 65 & Stadium co-location (edge) \\
\texttt{mec\_industrial\_1} & 15 & 12 & 3 & 25 & 40 & 30 & Factory proximity (edge) \\
\texttt{metro\_agg\_hub} & 8 & 5 & 10 & 15 & 25 & 50 & Metro aggregation (metro) \\
\texttt{regional\_dc\_north} & 35 & 30 & 35 & 40 & 50 & 40 & Regional data centre (core) \\
\bottomrule
\end{tabular}
\begin{flushleft}
\footnotesize
Note: Latency values represent end-to-end propagation delay from each RAN sector to the UPF node, reflecting geographic distance and network topology. Compute load indicates current resource utilisation percentage.
\end{flushleft}
\end{table*}

After each LLM call, the orchestrator's response is appended to the conversation history. The response is then parsed to detect action strings through pattern matching. If the response contains \texttt{ACTION: CALL\_AGENT}, the system extracts the agent name and request, invokes functions to consult specialist agents, and appends the specialist's response as an observation to the conversation history. If the response contains \texttt{ACTION: PROVISION\_SLICE}, the system extracts the configuration parameters and invokes functions to perform slice provisioning. If the response contains \texttt{ACTION: FINISH}, the loop terminates.

All LLM calls are made with temperature set to 0.0 to ensure deterministic reasoning and reproducibility across experimental runs. We use the Cerebras Cloud API with the Llama-3.3-70B model to obtain agents' responses. Token usage for each LLM call (both orchestrator and specialist invocations) is tracked and aggregated for computational efficiency analysis.

\section{Experiments}
\label{sec:experiments}

This section presents experimental validation of the proposed hierarchical agentic AI framework across multiple evaluation dimensions. We begin by describing the experimental setup, then assess system performance on the benchmark suite, conduct comparative analysis against baseline configurations, perform ablation studies to quantify component contributions, and identify critical prompt engineering challenges.

\subsection{Experimental Setup}

We evaluate system performance through a simulation-based framework in which the network state $\mathcal{S}$ is represented as a static structured object (as defined in Section~\ref{sec:sysmodel}) that remains fixed across all experimental runs, isolating the agentic system's semantic reasoning capabilities from the variability of live network conditions.

The benchmark suite comprises 12 intent-based network slice provisioning scenarios spanning the three foundational 6G use case categories defined in Section~\ref{sec:usecase}. The benchmark includes four URLLC scenarios addressing safety-critical applications such as e-sports tournaments, industrial automation, connected emergency vehicles, and augmented reality maintenance systems; four eMBB scenarios encompassing high-capacity streaming, suburban fixed wireless access, mobile video surveillance, and public Wi-Fi deployments; and four mMTC scenarios covering smart metering infrastructure, agricultural IoT networks, crowd analytics, and environmental monitoring systems.

The mock network topology models a representative metropolitan-scale 6G deployment. The radio access network comprises five sectors spanning diverse deployment contexts from high-density urban venues to rural wide-area coverage, as detailed in Table~\ref{tab:ran_sectors_topology}. The core network implements a three-tier compute architecture with sector-dependent latency characteristics, as detailed in Table~\ref{tab:upf_nodes_topology}.

Each benchmark scenario is executed three times to verify consistency, with results averaged across runs. The evaluation metrics (Semantic Accuracy and Engineering Utility) are computed as specified in Section~\ref{sec:evaluation}.

\begin{table*}[!t]
\caption{System Performance Across Benchmark Scenarios}
\label{tab:benchmark}
\centering
\begin{tabular}{llcccccc}
\toprule
\textbf{Scenario ID} & \textbf{Category} & \textbf{Semantic} & \textbf{Utility} & \textbf{Tokens} & \textbf{Time} & \textbf{Iterations} \\
& & \textbf{Accuracy} & \textbf{Score} & & \textbf{(s)} & \\
\hline
esports\_stadium              & URLLC  & 1.0 & 0.91 & 13363 & 3.2 & 3 \\
industrial\_automation        & URLLC  & 1.0 & 0.94 & 13573 & 4.0 & 3 \\
connected\_ambulance\_rural   & URLLC  & 1.0 & 0.54 & 13617 & 3.5 & 3 \\
ar\_maintenance\_industrial   & URLLC  & 0.5 & 0.94 & 12989 & 3.1 & 3 \\
stadium\_4k\_streaming        & eMBB   & 0.5 & 0.71 & 12587 & 3.3 & 3 \\
suburban\_fixed\_wireless     & eMBB   & 0.5 & 0.52 & 13731 & 3.3 & 3 \\
highway\_patrol\_video        & eMBB   & 0.5 & 0.62 & 13101 & 3.1 & 3 \\
plaza\_public\_wifi           & eMBB   & 0.5 & 0.71 & 13799 & 3.3 & 3 \\
smart\_meters\_suburban       & mMTC   & 0.5 & 0.52 & 13085 & 3.6 & 3 \\
agricultural\_iot\_rural      & mMTC   & 0.5 & 0.86 & 13021 & 3.2 & 3 \\
stadium\_crowd\_analytics     & mMTC   & 1.0 & 0.91 & 13557 & 3.5 & 3 \\
environmental\_monitoring     & mMTC   & 0.5 & 0.79 & 12951 & 3.1 & 3 \\
\hline
\textbf{Mean $\pm$ Std} & & \textbf{0.667 $\pm$ 0.246} & \textbf{0.747 $\pm$ 0.169} & \textbf{13281 $\pm$ 377} & \textbf{3.81 $\pm$ 1.82} & \textbf{3.0 $\pm$ 0.0} \\
\bottomrule
\end{tabular}
\end{table*}

\subsection{Performance on Benchmark Scenarios}

Table~\ref{tab:benchmark} presents system performance across the 12 benchmark scenarios. The proposed multi-agent architecture achieves a mean Semantic Accuracy of 0.667 (standard deviation 0.246), with four scenarios (33\%) receiving Perfect Match scores of 1.0 and seven scenarios (58\%) achieving Partial Match scores of 0.5. The mean Engineering Utility score of 0.747 (standard deviation 0.169) exceeds the predefined feasibility threshold of 0.70, demonstrating that provisioned configurations satisfy technical constraints.

Computational efficiency metrics reveal an average token consumption of 13,281 tokens per provisioning decision. The mean decision latency of 3.8 seconds positions the system within the operational timescale of non-real-time provisioning workflows, where network slice configurations are established prior to service activation rather than adapted during real-time operation. This latency is consistent with policy-level orchestration decisions that operate on the timescale of seconds to minutes, as distinct from per-packet or per-slot control requiring sub-millisecond responses. The system consistently converges within 3.0 ReAct iterations across all scenarios, demonstrating efficient reasoning without excessive computational overhead.

We observed a systematic bias from these results. Analysis of the benchmark results revealed a systematic pattern in decision-making: scenarios requiring regional data centre placement for delay-tolerant traffic exhibited lower semantic accuracy compared to ultra-low latency scenarios. Manual inspection of Core specialist reasoning traces identified the root cause as insufficiently precise policy encoding in the agent's system prompt. The Core specialist consistently selected the lowest-latency option available regardless of traffic type, failing to apply cost-efficiency policies for delay-tolerant applications such as eMBB streaming and mMTC sensor deployments.

The issue stemmed from weak directive language in the Core specialist prompt. The original instruction stated that regional data centres were ``acceptable'' for moderate latency requirements, which the language model interpreted as ``only use if nothing better exists'' rather than ``prefer for cost efficiency when latency tolerance permits.'' The prompt lacked explicit traffic-type awareness, failing to distinguish between URLLC applications where milliseconds matter for safety and eMBB or mMTC applications where regional placement is operationally equivalent but economically superior.

We revised the Core specialist prompt to encode traffic-aware placement policies more explicitly. The updated prompt instructs the agent to: (1) reserve edge nodes exclusively for URLLC traffic with safety-critical latency requirements, (2) prefer regional data centres for eMBB streaming applications where 30--50~ms latency is imperceptible to end users due to client-side buffering, and (3) mandate regional placement for mMTC deployments exhibiting high latency tolerance. The revised prompt replaces passive language (``acceptable'') with directive language (``prefer'', ``avoid'', ``mandate'') and includes worked examples demonstrating correct regional data centre selection for delay-tolerant traffic.

Following prompt revision, system performance improved from 0.667 to 0.708 semantic accuracy (6 to 7 scenarios achieving perfect match). Whilst this improvement is statistically modest, it demonstrates that the shift from permissive language (``acceptable'') to directive language (``prefer'', ``mandate'') altered decision logic across multiple scenarios, with the lexical change compounding across eMBB and mMTC categories where cost-efficient placement was previously overlooked.

This finding exposes a critical challenge for production deployment of agentic systems: how detailed should system prompts be to achieve reliable behaviour without incurring prohibitive computational costs or creating brittle over-specification? Overly concise prompts risk ambiguity and inconsistent interpretation, as demonstrated by the original ``acceptable'' formulation. Conversely, exhaustively detailed prompts increase token consumption (and thus inference cost), potentially introduce contradictory instructions, and may not generalise to more complex network topologies. The optimal prompt design strategy likely varies across domains, model capabilities, and deployment contexts, representing an open research question requiring systematic investigation.

\subsection{Qualitative Analysis: System Reasoning Behaviour}
\label{sec:qualitative}

\begin{table*}[!t]
\centering
\caption{Condensed Reasoning Trace for URLLC Industrial Automation Scenario}
\label{tab:reasoning_summary}
\small
\begin{tabular}{p{0.95\textwidth}}
\toprule
\textbf{Iteration 1: Intent Parsing and RAN Consultation} \\
\midrule
Orchestrator classifies traffic as URLLC based on keywords ``ultra-low latency'' (<5ms explicit threshold) and ``safety-critical operations'', identifying dual constraints of latency and reliability for industrial control application. \\[0.3em]
\texttt{ACTION: CALL\_AGENT | agent\_name=ran\_specialist | request=Can industrial\_park\_a support ultra-low latency and high reliability for automated robotic assembly with mmWave, mid-band, or low-band spectrum? What spectrum allocation is recommended?} \\
\midrule
\textbf{Iteration 2: RAN Response and Core Consultation} \\
\midrule
RAN specialist evaluates spectrum options considering reliability requirements in industrial environment (potential interference, penetration challenges). Recommends mid-band despite mmWave availability, justifying: \textit{``Mid-band offers better balance between capacity and penetration, crucial for maintaining high reliability in industrial setting with potential interference.''} Issues load monitoring warning (sector at 60\% utilisation). \\[0.3em]
Orchestrator synthesises RAN output and consults Core specialist, formulating URLLC-aware query emphasising stringent latency requirement: \\[0.3em]
\texttt{ACTION: CALL\_AGENT | agent\_name=core\_specialist | request=Given mid-band allocation at industrial\_park\_a for automated robotic assembly line requiring ultra-low latency (<5ms), which UPF node can provide necessary latency and what considerations for high reliability and current 60\% sector load?} \\
\midrule
\textbf{Iteration 3: Core Response and Configuration Synthesis} \\
\midrule
Core specialist evaluates latency matrix against strict 5ms threshold. Identifies mec\_industrial\_1 as sole compliant node (3ms latency), explicitly filtering options: mec\_stadium\_1 (15ms), metro\_agg\_hub (10ms), regional\_dc\_north (35ms) all exceed requirement. Validates compute capacity (30\% load, well below 85\% threshold). Justifies selection through application context: \textit{``Edge node dedicated to factory likely designed to handle specific interference and reliability challenges of industrial environment.''} \\[0.3em]
Orchestrator validates configuration consistency and provisions slice: mid-band @ industrial\_park\_a, UPF @ mec\_industrial\_1. \\
\bottomrule
\end{tabular}
\begin{flushleft}
\footnotesize
Note: Complete output available in Appendix~\ref{appendix:reasoning_trace}.
\end{flushleft}
\end{table*}

To contextualise the quantitative performance metrics presented above, we examine a representative execution trace illustrating the system's decision-making process. This qualitative analysis validates architectural claims regarding specialist coordination, iterative refinement, and prompt-engineered decision logic described in Section~\ref{sec:architecture}.

We consider the industrial automation scenario, which tests recognition of implicit URLLC requirements from safety-critical application context and appropriate edge node selection under latency constraints. The orchestrator receives the natural language intent: \textit{``Configure network slice for automated robotic assembly line at industrial\_park\_a. Requires ultra-low latency (<5ms) for real-time control and high reliability for safety-critical operations.''} The expert-defined Golden Standard specifies mid-band spectrum (balancing reliability and capacity for industrial environment) with edge node UPF placement (mandatory for sub-5ms latency requirement).

\subsubsection{Execution Summary}

The system converges in three iterations following the sequential coordination protocol. Table~\ref{tab:reasoning_summary} presents a condensed trace highlighting critical decision points, with complete system output provided in Appendix~\ref{appendix:reasoning_trace} for detailed examination.

\subsubsection{Analysis and Architectural Validation}

The execution trace validates three critical architectural properties. First, three-iteration convergence demonstrates efficient bounded reasoning (Algorithm~\ref{alg:react}) with sequential RAN-then-Core consultation ensuring proper constraint propagation. Second, the orchestrator exhibits intent parsing beyond keyword matching, inferring URLLC requirements from ``safety-critical operations'' and formulating traffic-type-aware queries that modulate specialist decision-making. The RAN specialist's recommendation of mid-band over mmWave, justified through reliability considerations in industrial environments, demonstrates domain-specific reasoning encoded through prompt engineering rather than procedural logic. Third, the Core specialist's systematic candidate filtering against the 5ms threshold validates strict enforcement characteristic of URLLC placement policies, contrasting with cost-optimising eMBB behaviours observed in streaming scenarios.

Evaluation against the Golden Standard yields Semantic Accuracy of 1.0 (Perfect Match) and Engineering Utility of 0.94, comprising $S_{\text{latency}}=1.0$, $S_{\text{resource}}=1.0$, and $S_{\text{congestion}}=0.40$. Execution metrics reveal 13,573 total tokens with 4 seconds decision latency. The structured reasoning traces enable operator audit of decision rationale, addressing explainability requirements for autonomous systems in regulated environments.

\subsection{Comparative Analysis Against Baseline Approaches}

\begin{figure}[!t]
    \centering
    \includegraphics[width=\columnwidth]{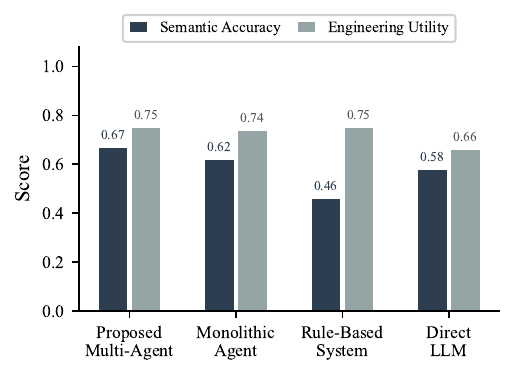}
    \caption{Comparative performance across baseline approaches.}
    \label{fig:comparative}
\end{figure}

To validate the contribution of the hierarchical multi-agent architecture, we compare the proposed system against three baseline configurations representing alternative design choices. Table~\ref{tab:comparison} and Fig.~\ref{fig:comparative} presents comparative performance metrics across the benchmark suite.

The proposed multi-agent system achieves an 8\% improvement in Semantic Accuracy over the monolithic single-agent baseline (0.67 versus 0.62). The monolithic agent, despite receiving identical network state information and tool access, demonstrates higher decision variance, suggesting that the increased cognitive load of reasoning over both RAN and Core domains simultaneously leads to less consistent decision-making. However, the monolithic baseline achieves 86\% token reduction (1,894 versus 13,281 tokens) and 79\% latency reduction (0.8s versus 3.8s), highlighting a fundamental trade-off between decision quality and computational efficiency.

The rule-based system demonstrates competitive Engineering Utility scores (0.75) approaching the multi-agent system's performance, reflecting strict adherence to predefined engineering constraints. However, the rule-based approach achieves only 0.46 Semantic Accuracy, representing a 31\% degradation relative to the multi-agent system. Manual inspection reveals that the rule-based system struggles with natural language variations in intent expression: it successfully handles explicit keywords (``ultra-low latency'') but fails to recognise synonymous expressions (``real-time responsiveness'', ``instantaneous connectivity''). This brittleness to linguistic variation underscores the value of LLM-based semantic understanding, which generalises across paraphrases without explicit rule authoring.

\begin{table}[!t]
\caption{Comparative Performance Against Baseline Approaches}
\label{tab:comparison}
\centering
\begin{tabular}{lcccc}
\toprule
\textbf{System} & \textbf{Semantic} & \textbf{Utility} & \textbf{Tokens} & \textbf{Time} \\
\textbf{Configuration} & \textbf{Accuracy} & \textbf{Score} & & \textbf{(s)} \\
\hline
Proposed Multi-Agent & 0.67 & 0.75 & 13281 & 3.8 \\
Monolithic Agent     & 0.62 & 0.74 & 1894  & 0.8 \\
Rule-Based System    & 0.46 & 0.75 & N/A   & 0.1 \\
Direct LLM           & 0.58 & 0.66 & 1157  & 0.6 \\
\bottomrule
\end{tabular}
\end{table}

The direct LLM prompting baseline, which queries a single LLM instance without iterative refinement or specialist consultation, achieves the lowest performance (0.58 Semantic Accuracy, 0.66 Utility). Whilst this configuration offers superior computational efficiency (1,157 tokens, 0.6 seconds), the lack of structured reasoning leads to premature decisions that frequently violate feasibility constraints. This finding aligns with prior work demonstrating the benefits of iterative reasoning patterns in LLM-based systems \cite{yao2023react}.

\subsection{Ablation Study Results}

\begin{figure}[!t]
    \centering
    \includegraphics[width=\columnwidth]{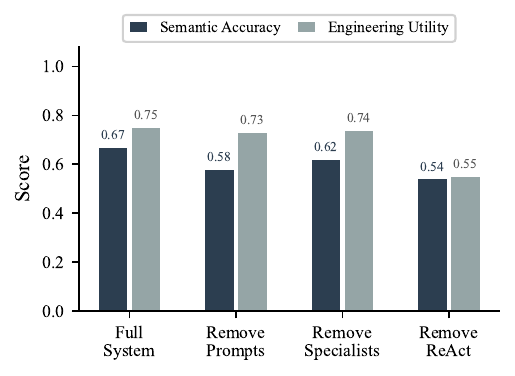}
    \caption{Ablation study results across architectural components.}
    \label{fig:ablation}
\end{figure}

To quantify the contribution of individual architectural components, we conduct systematic ablation experiments in which specific elements are disabled or modified. Table~\ref{tab:ablation} and Fig.~\ref{fig:ablation} presents performance degradation relative to the full system configuration.

\textbf{Remove Engineered System Prompts.} Replacing detailed specialist prompts with minimal generic prompts (``You are a helpful AI assistant specialising in 6G networks...'') whilst retaining network state access results in moderate degradation ($\Delta$ Semantic Accuracy = $-0.09$). This ablation tests whether carefully encoded decision rules (e.g., ``>10k users requires mmWave'', ``load >80\% triggers warning'', ``prefer regional DC for eMBB traffic'') meaningfully improve performance beyond the base LLM's general 6G knowledge. The modest degradation suggests that the base Llama-3.3-70B model possesses substantial general telecommunications knowledge but benefits from structured prompts that encode quantitative thresholds and operational policies. Without explicit decision rules, the system exhibited inconsistent threshold application and missed cost-optimisation opportunities, though it maintained reasonable baseline performance (0.58 accuracy) through general-purpose reasoning.

\begin{table}[!t]
\caption{Ablation Study: Component Contribution Analysis}
\label{tab:ablation}
\centering
\begin{tabular}{lcccc}
\toprule
\textbf{Configuration} & \textbf{Semantic} & \textbf{Utility} & \textbf{$\Delta$ Sem} & \textbf{$\Delta$ Util} \\
& \textbf{Accuracy} & \textbf{Score} & & \\
\hline
Full System         & 0.67 & 0.75 &  0.00 &  0.00 \\
Remove Prompts      & 0.58 & 0.73 & -0.09 & -0.02 \\
Remove Specialists  & 0.62 & 0.74 & -0.05 & -0.01 \\
Remove ReAct        & 0.54 & 0.55 & -0.13 & -0.20 \\
\bottomrule
\end{tabular}
\end{table}

\textbf{Remove Specialist Agents.} Eliminating hierarchical specialist consultation and configuring the orchestrator to make all decisions monolithically reduces Semantic Accuracy by 0.05. This ablation validates that domain-specific expertise decomposition provides measurable improvement, though the effect is smaller than anticipated. The limited degradation suggests that for the tested scenarios (which involve relatively independent RAN and Core decisions), the primary benefit of specialist agents is cognitive load reduction rather than fundamentally superior reasoning. More complex scenarios requiring joint optimisation across domains may exhibit larger degradation.

\textbf{Remove ReAct Iteration.} Disabling the iterative refinement loop and forcing single-pass generation results in the largest degradation ($\Delta$ Semantic Accuracy = $-0.13$, $\Delta$ Utility = $-0.20$). Without iteration, the orchestrator cannot revise decisions based on specialist feedback, recover from initial parsing errors, or refine configurations through multi-turn reasoning. The substantial utility degradation (from 0.75 to 0.55) indicates that single-pass decisions frequently violate engineering constraints that would have been caught and corrected in subsequent iterations. This finding confirms that the ReAct reasoning pattern contributes meaningfully to both semantic understanding and technical feasibility.

\section{Discussions}
\label{sec:discussion}

The experimental results provide several insights into agentic AI system design for network automation. First, the identification and correction of the Core specialist's ``latency greedy'' behaviour demonstrates that LLM-based systems can exhibit subtle prompt-induced biases that manifest as systematic failure patterns rather than random errors. Unlike traditional software bugs that cause immediate crashes, prompt engineering issues create behavioural tendencies that compound across scenarios, requiring careful error pattern analysis to diagnose. The improvement from prompt revision (0.667 to 0.708 semantic accuracy) validates prompt engineering as a critical architectural design parameter requiring rigorous validation.

Second, the comparative analysis reveals fundamental trade-offs between decision quality and computational efficiency. The multi-agent system achieves superior semantic understanding (+8\% over monolithic) but requires 7$\times$ more tokens and 5$\times$ longer execution time. For production deployments, this suggests a tiered architecture: complex, ambiguous intents processed by the full multi-agent system; straightforward intents handled by lightweight monolithic agents; and repetitive patterns addressed by cached rule-based configurations. Such hybrid systems could optimise the accuracy-efficiency Pareto frontier.

Third, the ablation studies indicate that ReAct iterative refinement provides the largest marginal contribution, followed by engineered system prompts and specialist decomposition. This ordering suggests that architectural efforts should prioritise robust reasoning protocols and precise prompt engineering over organisational structure. However, these findings are specific to the tested scenario characteristics (largely decoupled RAN/Core decisions, moderate complexity intents) and may not generalise to more sophisticated IBN workflows requiring tighter cross-domain coordination.

The sequential coordination protocol described in Section~\ref{sec:coordination} does not currently include a backtracking mechanism whereby the Core specialist can request a revised RAN recommendation when no UPF node satisfies the latency constraint for the selected sector. In the tested scenarios, this limitation is bounded by two factors. Sector selection is primarily driven by geographic context specified in the intent rather than by downstream UPF availability, and the orchestrator performs a final consistency check that identifies cross-domain conflicts before provisioning. Nevertheless, scenarios requiring joint optimisation, such as RAN congestion necessitating sector migration with consequential UPF re-selection, would benefit from an iterative negotiation loop between specialists.

The token efficiency comparison (proposed 13,281 versus monolithic 1,894) highlights a critical challenge for LLM-based network automation: inference costs scale with architectural complexity. At current pricing (1 USD per million tokens), the proposed system costs approximately 0.013 USD per provisioning decision. Future work should investigate prompt compression techniques, knowledge distillation to smaller models, and caching strategies to reduce per-decision token consumption whilst preserving reasoning quality.

Regarding the simplified network representation used in this study, we note that whilst the experimental evaluation employs mock network data for controlled validation, the architectural principles and reasoning patterns are designed to generalise to production environments. Future work should investigate integration with actual network management platforms and evaluate performance under dynamic network conditions. Specifically, O-RAN environments represent a promising deployment target, where the orchestrator agent could function as a Non-Real-Time RAN Intelligent Controller (Non-RT RIC) application coordinating xApps and rApps configurations based on natural language operator intents, interfacing with E2 telemetry interfaces for network state acquisition. The concept of representing network state as structured data accessible to agents is more fundamental than the specific implementation approach, and similar representations could be derived from network management systems, operational support systems, or O-RAN RIC frameworks in production deployments.

\section{Conclusion}
\label{sec:conclusion}

This paper presented a hierarchical multi-agent framework for IBN in 6G environments, demonstrating that LLM-powered agentic systems can autonomously translate natural language operational intents into executable network slice configurations. The architecture employs an orchestrator agent coordinating two domain-specific specialist agents through iterative ReAct reasoning, with all agents grounded in structured network state representations. Experimental evaluation across diverse benchmark scenarios showed that the proposed system outperforms monolithic agents, rule-based systems, and direct LLM prompting, with ablation studies confirming that iterative refinement and prompt engineering provide the most significant architectural contributions. As 6G networks and O-RAN architectures evolve towards autonomous operation, agentic AI frameworks offer a promising path forward for transforming natural language intents into network-wide configurations whilst maintaining transparency and operational trust through interpretable reasoning traces.

\bibliographystyle{IEEEtran}
\bibliography{bare_jrnl}

\newpage
\onecolumn

\appendix

\subsection{Complete Reasoning Trace for URLLC Industrial Automation Scenario}
\label{appendix:reasoning_trace}

This trace corresponds to the qualitative analysis presented in Section~\ref{sec:qualitative}. The scenario tests recognition of implicit URLLC requirements from safety-critical application context, appropriate spectrum selection balancing reliability and capacity in industrial environments, and strict latency threshold enforcement for edge node placement.

\textit{1) Input Intent}

\begin{mdframed}[style=promptbox, frametitle={\textbf{User Intent}}]
\begin{lstlisting}
Configure network slice for automated robotic assembly line at industrial_park_a. Requires ultra-low latency (<5ms) for real-time control and high reliability for safety-critical operations.
\end{lstlisting}
\end{mdframed}

\textit{2) Golden Standard Configuration}

\begin{itemize}
\item \textbf{Sector:} industrial\_park\_a
\item \textbf{Spectrum:} mid\_band (balancing reliability and capacity for industrial environment with potential interference)
\item \textbf{UPF Node:} mec\_industrial\_1 (edge node achieving 3ms latency, satisfying <5ms requirement)
\item \textbf{Rationale:} Ultra-low latency requirement (<5ms) mandates edge placement. Safety-critical operations require high reliability, favouring mid-band over mmWave due to superior propagation characteristics in industrial environments with obstacles and interference. Industrial control applications exhibit moderate bandwidth requirements (500 users), making mid-band capacity sufficient.
\end{itemize}

\textit{3) Network State Context}

The industrial\_park\_a sector exhibits the following characteristics:
\begin{itemize}
\item Active users: 500 (moderate density)
\item Load percentage: 60\% (elevated but manageable)
\item Spectrum availability: 200 MHz mmWave, 100 MHz mid-band, 10 MHz low-band
\item Environmental context: Factory zone with high interference, critical reliability requirements
\end{itemize}

Latency matrix for industrial\_park\_a:
\begin{itemize}
\item mec\_stadium\_1: 15ms (edge, geographically distant)
\item mec\_industrial\_1: 3ms (edge, co-located with factory)
\item metro\_agg\_hub: 10ms (metro tier)
\item regional\_dc\_north: 35ms (regional tier)
\end{itemize}

\textit{4) Complete Execution Trace}
\begin{mdframed}[style=promptbox, frametitle={\textbf{Complete Execution Trace}}]
\begin{lstlisting}
--- Iteration 1 ---
[Thinking]: THOUGHT: The intent requires ultra-low latency (<5ms) and high reliability for safety-critical operations in an automated robotic assembly line at industrial_park_a. This implies a need for a low-latency and highly reliable network configuration. To achieve this, I need to consult the RAN specialist for spectrum allocation and the Core specialist for UPF placement. First, I will consult the RAN specialist to determine the appropriate spectrum for industrial_park_a that can support the required reliability and capacity for the automated robotic assembly line.

ACTION: CALL_AGENT | agent_name=ran_specialist | request=Can industrial_park_a support ultra-low latency and high reliability for automated robotic assembly with mmWave, mid-band, or low-band spectrum? What spectrum allocation is recommended?

   -> Calling ran_specialist...
   <- Agent Response: **ANALYSIS FOR INDUSTRIAL_PARK_A**

1. **User Density**: 500 active users, which is considered moderate density.
2. **Application Requirements**: Ultra-low latency and high reliability for automated robotic assembly.
3. **Load Status**: 60% load, which is relatively high but not at the congestion threshold.
4. **Spectrum Availability**:
   - mmWave: 200 MHz available
   - mid-band: 100 MHz available
   - low-band: 10 MHz available

**DECISION PROCESS**

- **Ultra-Low Latency Requirement**: This suggests a preference for mmWave due to its potential for very low latency and high capacity. However, the availability of mmWave is limited to 200 MHz.
- **High Reliability Requirement**: For critical applications like automated robotic assembly, reliability is paramount. mmWave offers high capacity but may suffer from interference and penetration issues. Mid-band offers a better balance of capacity and penetration, potentially enhancing reliability.
- **Moderate Density**: The number of active users is moderate, suggesting that mid-band could be sufficient in terms of capacity, especially considering the need for reliability.

**RECOMMENDATION**

Given the requirements for ultra-low latency and high reliability, and considering the moderate user density and available spectrum, **mid-band** is recommended for industrial_park_a. Although mmWave is typically preferred for ultra-low latency applications due to its high capacity, the limited availability of mmWave (200 MHz) and the critical need for reliability in this scenario make mid-band a more appropriate choice. Mid-band offers a better balance between capacity and penetration, which is crucial for maintaining high reliability in an industrial setting with potential interference.

Mid-band's 100 MHz availability should be sufficient for the current 500 active users, especially if the application's bandwidth requirements are managed efficiently. Additionally, using mid-band helps in mitigating potential issues with mmWave's limited range and penetration, which could be problematic in an industrial environment with various obstacles.

**WARNING**

While the current load of 60% is manageable, it is essential to monitor the sector's load closely to avoid approaching the congestion threshold. As the load increases, reassessing spectrum allocation and considering upgrades or redistribution of users across different bands may be necessary to maintain the required ultra-low latency and high reliability.

**OUTPUT**

```
RECOMMENDATION: Use mid-band at industrial_park_a because it offers a balance of capacity and reliability necessary for ultra-low latency applications like automated robotic assembly, given the moderate user density and available spectrum.
WARNING: Sector industrial_park_a is at 60% load; monitor closely to avoid congestion and consider spectrum reallocation or upgrades as necessary.
```...

--- Iteration 2 ---
[Thinking]: THOUGHT: The RAN specialist recommends using mid-band at industrial_park_a for the automated robotic assembly line due to its balance of capacity and reliability, which is crucial for ultra-low latency and safety-critical operations. The specialist also warns about the sector being at 60% load, suggesting the need for close monitoring to avoid congestion. Next, I need to consult the Core specialist to determine the optimal UPF placement for achieving ultra-low latency (<5ms) given the mid-band spectrum allocation at industrial_park_a.

ACTION: CALL_AGENT | agent_name=core_specialist | request=Given mid-band allocation at industrial_park_a for an automated robotic assembly line requiring ultra-low latency (<5ms), which UPF node can provide the necessary latency and what considerations should be taken for high reliability and the current 60% sector load?    

   -> Calling core_specialist...
   <- Agent Response: To address the orchestrator's request for ultra-low latency (<5ms) at industrial_park_a for an automated robotic assembly line, we need to consider the latency matrix and the current network state.

1. **Identify the target sector**: industrial_park_a
2. **Look up latencies** in the `latency_matrix` for industrial_park_a:
   - mec_stadium_1: 15ms (too high)
   - mec_industrial_1: 3ms (meets the requirement)
   - metro_agg_hub: 10ms (too high)
   - regional_dc_north: 35ms (too high)
3. **Extract the latency requirement**: <5ms
4. **Filter nodes** that meet the latency requirement: mec_industrial_1 is the only node that meets this requirement.
5. **Check compute capacity** for the candidate node: mec_industrial_1 has a compute_load_percent of 30%, which is well below the 85% threshold, indicating ample capacity.
6. **Considerations for high reliability**: Given the critical nature of the application (automated robotic assembly line) and the sector's description ("Factory zone. High interference, critical reliability."), it's essential to prioritize reliability. mec_industrial_1, being an edge node dedicated to the factory, is likely designed to handle the specific interference and reliability challenges of the industrial environment.
7. **Current sector load**: The sector load at industrial_park_a is 60%, which is manageable. Deploying the UPF at mec_industrial_1 should not significantly impact the sector's load, given the node's current compute load percent.

**RECOMMENDATION**: Deploy UPF at mec_industrial_1 to achieve 3ms latency from industrial_park_a.
This recommendation meets the ultra-low latency requirement, considers the reliability needs of the application, and takes into account the current network state and sector load....

--- Iteration 3 ---
[Thinking]: THOUGHT: The Core specialist recommends deploying the UPF at mec_industrial_1 to achieve a latency of 3ms, which meets the ultra-low latency requirement of <5ms for the automated robotic assembly line at industrial_park_a. This recommendation also considers the reliability needs of the application and the current network state, including the sector load. With both RAN and Core specialist recommendations in hand, I can now provision the network slice.

ACTION: PROVISION_SLICE | slice_id=industrial_autonomy_001 | ran_config=mid-band@industrial_park_a | core_config=UPF@mec_industrial_1
ACTION: FINISH | summary=Network slice industrial_autonomy_001 configured for automated robotic assembly line at industrial_park_a. RAN: mid-band for reliability and capacity. Core: UPF at mec_industrial_1 for 3ms latency. Sector load at 60%; monitoring recommended to maintain performance.
\end{lstlisting}
\end{mdframed}

\textit{5) Evaluation Against Golden Standard}

\textbf{Generated Configuration:}
\begin{itemize}
\item Sector: industrial\_park\_a
\item Spectrum: mid\_band
\item UPF Node: mec\_industrial\_1
\end{itemize}

\textbf{Golden Standard:}
\begin{itemize}
\item Sector: industrial\_park\_a
\item Spectrum: mid\_band
\item UPF Node: mec\_industrial\_1
\end{itemize}

\textbf{Semantic Accuracy:} 1.0 (Perfect Match - all three dimensions correct)

\textbf{Engineering Utility Calculation:}
\begin{itemize}
\item $S_{\text{latency}} = 1.0$ (edge node: 3ms latency <10ms threshold)
\item $S_{\text{resource}} = 1.0$ (mid-band appropriate for industrial control bandwidth)
\item $S_{\text{congestion}} = 0.40$ (sector load 60\%)
\item Weights for URLLC: $w_l = 0.8$, $w_r = 0.1$, $w_c = 0.1$
\item $U = 0.8 \times 1.0 + 0.1 \times 1.0 + 0.1 \times 0.40 = 0.94$
\end{itemize}

\textbf{Execution Metrics:}
\begin{itemize}
\item Total iterations: 3
\item Total tokens: 13,573
\item Decision latency: 4 seconds
\item Specialist consultations: 2 (RAN: 1, Core: 1)
\end{itemize}

\end{document}